\documentclass{ws-ijicc}

\begin{document}

%
\catchline{}{}{}{}{}
%

\markboth{Hostettler}
{Organizational Buying Behavior and HRI}

\title{Using vs. Purchasing Industrial Robots:\\Adding an Organizational Perspective to Industrial HRI
}

\author{DAMIAN HOSTETTLER}

\address{Institute of Computer Science, University of St. Gallen, Switzerland, Rosenbergstrasse 30, 9000 St. Gallen, {damian.hostettler\@student.unisg.ch}}

\maketitle

\begin{abstract}
\begin{itemize}
    \item Purpose: Industrial robots allow manufacturing companies to increase productivity and remain competitive. For robots to be used, they must be accepted by operators on the one hand and bought by decision-makers on the other. The roles involved in such organizational processes have very different perspectives. It is therefore essential for suppliers and robot customers to understand these motives so that robots can successfully be integrated on manufacturing shopfloors.
    \item Methodology: We present findings of a qualitative study with operators and decision-makers from two Swiss manufacturing SMEs. Using laddering interviews and means-end analysis, we compare operators' and deciders' relevant elements and how these elements are linked to each other on different abstraction levels. These findings represent drivers and barriers to the acquisition, integration and acceptance of robots in the industry.
    \item Findings: We present the differing foci of operators and deciders, and how they can be used by demanders as well as suppliers of robots to achieve robot acceptance and deployment. First, we present a list of relevant attributes, consequences and values that constitute robot acceptance and/or rejection. Second, we provide quantified relevancies for these elements, and how they differ between operators and deciders. And third, we demonstrate how the elements are linked with each other on different abstraction levels, and how these links differ between the two groups.
    \item Practical implications: Our findings are beneficial to suppliers as well as customers of industrial robots on several levels. Regarding robot deployment and acceptance, they inform suppliers what they need to focus on to increase customers' willingness to buy robots while supporting deciders in their challenges to convince operators how robots support them in reaching their individual goals. Moreover, our classification of customer needs on different abstraction levels allows suppliers as well as demanders to align not only product attributes and customer needs, but also value conceptions on strategic and cultural levels.
    \item Originality: Existing HRI research focuses on technology or psychological aspects, but does not sufficiently consider the organizational perspective. Our research and findings fill this gap by providing detailed insights into the relevant connections between product attributes and use consequences that arise from the individual values of operators and decision-makers. We provide findings on buying center structures in SMEs as well as on operators' and deciders' motives that support the acceptance and thus the deployment of robots, thereby strengthening the strategies of robot suppliers as well as the competitiveness of manufacturing companies.
\end{itemize}
\end{abstract}


\section{Introduction}\label{sec1}

Industrial manufacturing shopfloors have undergone several transitions in recent decades and the trend towards smart factories will likely continue~\citep{Mabkhot2018}. To realize smart factories, robots play a key role; nowadays, they are no longer standalone devices that work in isolation but increasingly become team members in production facilities. Especially with the rise of collaborative robots (cobots) that need fewer safety precautions, closer human-robot collaboration (HRC) where agency is distributed among humans and robots becomes reality~\citep{Weiss2021CobotsCollaboration}. With the global robot density in manufacturing nearly doubling from 2015 to 2020~\citep{IFR2021ExecutiveRobots}, resulting in 3.9 million operational robots in 2022~\citep{IFR2024} and sales by major cobot manufacturer Universal Robots (UR) reaching more than 75'000 cobots sold in 2023\footnote{\url{https://www.universal-robots.com/media/1828242/07\_2023\_ur\_media\_kit.pdf}}, this topic will likely keep gaining relevance. 
Over the past decades, the scientific community has put considerable effort into achieving a better understanding of human-robot interactions (HRI). While most HRI studies focus on social or humanoid robots~\citep{Kopp2020}, researchers have started to investigate the acceptance of \emph{industrial} robots with actual factory workers~\citep{Elprama2016AcceptanceRobots, Meissner}. Furthermore, drivers and barriers of implementing cobots have received only little attention; and, if so, only few studies include insights from actual industry agents: \citet{Cigdem} studies the effect of attitudes toward robots and trust in human–robot cooperation on the intention to use industrial robots with Turkish factory workers. Their results confirm that negative attitudes as well as trust significantly affects intention to use. \citet{KILDAL201821} compares actual robot operators' expectations with those of students focusing on user experience (UX), and found that safety, usability, and flexibility are most relevant for both groups, even though the relevance of safety is higher for professionals, whereas voice and gesture-based interactions were more favored by students. \citet{AALTONEN20191151} compares industry's and academia's experiences and expectations and found lack of knowledge to be the most significant barrier for cobot adoption. \citet{CORREIASIMOES2020101574} presents internal and external factors that influence managers to use cobots from an adoption point of view. Internal factors such as the receptiveness of an organization toward innovations, the risk-taking climate and the organization's readiness, and external factors such as pressure from competitors and customers were found to influence the adoption of cobots.
To continue improving human-robot collaboration,~\citet{Weiss2021CobotsCollaboration} supports that we need inputs ``from various scientific disciplines, including robotics, design, psychology, sociology, and so on'', ``but also from affected stakeholders (operators, maintainers, shift leads, etc.)''. 
In addition to this variety of topics that promote or hinder the existence of robots in shopfloors, we observe one major difference between most humanoid robots and industrial robots that is already visible when it comes to the \emph{purchasing} decision: Industrial robots are purchased by companies rather than by individuals, and the purchasing decision hence does not only depend on individual preference, but on motives that are linked to the individual's role and context in that company. We therefore propose to consider the purchasing process (from needs assessment to daily usage) as well as different stakeholders (such as management boards, plant managers or operators) to enrich the creation of positive interactions between humans and industrial robots. 
To overcome the current situation where most of the existing research addresses the topics of robot acceptance from a purely robot-oriented point of view~\citep{Prati2021}, \citet{Weiss2021CobotsCollaboration} calls for inclusion of human-related and contextual factors to enable human trust in robots. Accordingly, the organizational context with its respective actors enhances industrial HRI research and helps to understand \emph{who} is involved \emph{in which phase} of the purchase process, and what the respective \emph{motives} of these stakeholder groups are. To address this, the present study investigates drivers and barriers of the acceptance of industrial robots and cobots along the purchasing process, where we take into account the needs and reservations of individuals regarding their specific roles (e.g., individuals who are involved in the purchase decision vs. operators who work directly with the robot) and contexts (e.g., the purchase decision vs. day-to-day usage of a robot).
We propose that the renowned concept of \emph{Organizational Buying Behavior (OBB)}~\citep{Webster} provides us with a lens that permits to investigate the above-mentioned roles in an organizational context. However, a recent literature review confirms that only 5-10\% of premier marketing journals focus on Business-to-Business (B2B) settings and that the number of Buying Center (BC) manuscripts has decreased since the late 2000s~\citep{CABANELAS202365}. In addition, OBB and BCs in small and medium enterprises (SMEs) are characterized by few BC members and limited resources, compared to larger companies, and are underrepresented in current research~\cite{Cardinali}, despite their economic relevance. We therefore argue that taking into account how robots find their way onto manufacturing shopfloors in SMEs and understanding who is involved as well as the motives of these stakeholders are important aspects when investigating the acceptance and deployment of industrial robots. However, to the best of our knowledge, this has received no attention in neither published HRI nor marketing research so far.

\section{Related Work}\label{sec2}
To better understand the relevant relations between robots, human acceptance, and subsequent behavior with regard to the purchase and usage of robots on manufacturing shopfloors, we provide a brief overview of current research on the acceptance of robots and cobots, the effect of individual differences with regard to OBB, and on different HRI contexts.

\subsection{Robot Acceptance and Individual Differences}\label{acceptance}
Many research fields have examined the acceptance of technology and its relationship to human behavior. For example, the attribution of human behavior and even emotions to technical devices has been articulated in \emph{media equation theory} as early as 1996~\citep{Reeves1996TheMedia}. On this basis, several frameworks to measure technology and robot acceptance have been developed~\citep{Venkatesh2003UserView,Heerink2010AssessingModel,Weiss2009TheInteraction}. This is highly relevant for the future development of the robotics market across diverse fields. For instance, with respect to the hospitality industry, a recent study found that managers as well as front-line employees experience robot-phobia which causes stress and feelings of job insecurity, resulting in higher turnover intention, thereby intensifying labor shortage in the industry~\cite{Chen}. Better understanding of the \emph{organizational} perspective and the respective attitudes of decision-makers \emph{as well as} operators might permit overcoming such challenges.
\noindent
Focusing on robot acceptance, a large number of findings in HRI promote the idea that robot characteristics might be adapted to increase human acceptance. In a recent meta-analysis,~\citet{Esterwood2022} surveys several studies that demonstrate significant relationships of robot personality and human acceptance. Focusing on industrial robots, \citet{Hostettler2022} demonstrated effects of different robot movements on human preference. However, reliable findings on moderators that explain acceptance based on human characteristics are still scarce. Some studies revealed gender differences with respect to anthropomorphic and robotic movements~\citep{Abel2020GenderActions} and education differences, demonstrating, for example, that engineering students evaluate robots more favorably than psychology students~\citep{Szczepanowski2020EducationRobots}. Furthermore, gender, age, and prior experience with robots have been shown to influence individuals' attitude towards robots~\citep{Dinet2014ExploratoryUsers, Ivanov2018ConsumersEstablishmentse, Muller-Abdelrazeq2019InteractingContexts}. These findings indicate that individual differences have the potential to explain differing evaluations of HRI, and this understanding might then be used to increase robot acceptance. Regarding the organizational context, the individuals' acceptance of robots affects a company's willingness to buy, but also a supplier's selling effectiveness. Similar to the influence of individual differences in HRI, personal characteristics of purchasers in B2B contexts have been ``identified as a critical element in understanding and optimizing the buyer–seller relationship'', even though current research faces a lack of corresponding data availability~\cite{Mier}. In addition, comparing countries' robot density as a measure of worldwide automation state, the International Federation of Robotics' (IFR) recent World Robotic Report 2023 \citep{IFR2024} reveals a global average robot density of 151 robots per 10'000 employees. However, their numbers also demonstrate enormous differences between countries' adoption rates of industrial robots. While the Republic of Korea with 1'012 robots per 10'000 employees is the world's leading industrial robot adopter, Germany ranks third with 415 robots and the US ranks tenth with 285 robots per 10'000 employees. Accordingly, acceptance and adoption of industrial robots is crucial for countries to stay competitive in the global economy.
With regard to industrial cobots, some studies investigate the relationship between humans and their robotic colleagues in HRI. In a recent study,~\citet{Kopp2020} reviewed relevant literature and conducted a survey with German manufacturing companies on factors that facilitate or hinder the introduction of cobots. Their framework classifies relevant factors with regard to temporal phases (decision, implementation, and operation phase), and components of the socio-technical HRI system that comprises the cobot, human operator, working system and enterprise, and contextual factors. The survey reveals, for instance, that the technology-related factors \emph{occupational safety} and \emph{appropriate cobot configuration}, as well as the employee-centered factors \emph{fear of job loss} and \emph{ensuring an appropriate level of trust in the robot} were rated as most important~\citep{Kopp2020}.

These research streams allow us to understand technology acceptance in general as well as human and robot characteristics that promote acceptance. However, the perspective of organizations and their actors that represent actual industrial robot customers have so far been ignored in HRI research. We argue that enriching existing findings with these additional perspectives is highly relevant for both, the fields of HRI and industrial marketing alike.

\subsection{Organizational Buying Behavior}
\label{sec:orgbuy}

Research on OBB and its implications began more than 50 years ago and the core idea is still current today. Unlike everyday purchase decisions in Business-to-Consumer (B2C) markets, organizational buying involves ``many persons, multiple goals and potentially conflicting decision criteria'' \citep{Webster}. In addition to individuals' considerations and goals, interactions between involved persons and the fact that the individual's organization is exposed to various environmental influences need to be taken into account. Moreover, decision-making in organizations includes \emph{task} and \emph{non-task} motives. While the task dimension reflects ``the specific buying problem to be solved'', the non-task dimension includes the individual's ``achievement and risk-reduction motives''~\citep{Webster}. 
Furthermore, \emph{organizational roles} involved in buying processes reinforce the diversity of motives that influence the buying decision, indicating that findings presented in \citep{Kopp2020} might underlie additional complexity: When buying a robot, the CEO or owner of a company might want to increase efficiency as a consequence of pricing pressure, enabling their company to stay competitive and to gain experience with promising technologies on future shopfloors. Shopfloor workers in charge of defining the new robot's features might however have an incentive to badmouth the robot's abilities due to a fear of job loss. The workshop supervisor, on the other hand, might evaluate a robot's abilities differently, depending on their relationship to the worker whose job is at risk. And the CFO might decide for the cheapest product in order to realize a short-term incentive such as a bonus, with lower consideration for the workers' preferences, safe robot collaboration, or long-term strategic goals of the company. Accordingly, these differing motives help to understand how robots find its way into manufacturing shopfloors, and we propose to investigate diverse actors' motivations behind industrial robot purchase and usage. Even though robots are sold to companies, it is ultimately an individual or a group of individuals who decide whether to buy or not to buy a product~\cite{Mier}, and with regard to the shift from BCs to buying ecosystems, gaining a deep understanding of roles involved in purchase decisions for industrial robots might help to understand organizational roles as ``agile homo agens'', rather than homo economicus which might be outdated in this newer paradigm~\citep{EHRET2024}. In addition, prior research has highlighted that key influence of actual users in the decision-making process \cite{Howard}. As BCs in SMEs are typically smaller and purchasing processes differ from those in larger organizations, the relevance of the user as a ``center in its own''~\citep{EHRET2024} might be even higher in this context---however, regarding BC members, these users and their influence on purchasing processes are underrepresented in OBB research~\cite{Pedeliento}. In addition, understanding how an individual's personal characteristics affect sales effectiveness requires not only knowledge about the BC structure, but also the individual's characteristics beyond demographics---these are insufficient to explain buying decisions, as shown in~\cite{Mier}. Following current trends in procurement and B2B selling \cite{Bilro}, both relationship quality as well as value co-creation require a deep understanding of customers' buying centers as well as the involved individuals' needs. Achieving a detailed understanding of the involved individuals and their views does not only support actual robot deployment, thereby helping to overcome current challenges in industrial marketing. In addition, findings on actual actors' motives represent a relevant step towards operationalizing real-life buying centers which is required to close a relevant research gap in the marketing literature~\citep{CABANELAS202365}.

\subsection{HRI across Different Contexts}\label{context}

Regarding the context of HRI, several applications and environments have been investigated. However, most studies focus on very specific contexts and application scenarios are often criticized to be unrealistic \citep{Kopp2020}. For instance with regards to service robots, elderly care received particular consideration \citep{Sparrow2006,BEMELMANS2012114}. Furthermore, \citep{Joosse2013WhatPersonality} compares the two contexts of a robot being a cleaner or a tour guide and found that ``attraction  rules  for  robot  personalities  and  behaviors  depend  on  the  task  context''. In addition, cultural contexts result in differing HRI experiences~\citep{Salem2014}, and therefore need to be taken into account as well. Relating to manufacturing SMEs, organizations that deploy industrial robots generally focus on productivity gains, machine wear, or employee satisfaction. These specific contextual influences in HRI for instance need to be included when applying findings on movement parameters as proposed in \citet{Hostettler2022} when for example determining optimal robot speed, representing an entirely different context than B2C scenarios.

To include the organizational perspective and complement existing findings, we conducted a qualitative study focusing on the two contexts of \emph{buying} and \emph{using} industrial robots and cobots in Swiss manufacturing SMEs, and on the two major roles of \emph{decider} in the purchase decision and \emph{operator} of industrial robots. The following Section \ref{sec3} sets out the study design and the methods used.

\section{Method}\label{sec3}

The goal of this study is to investigate drivers and barriers of industrial robots in Swiss manufacturing SMEs and to compare them with regard to two roles involved in robot purchase or usage, namely deciders (management and board members) and operators. We conducted in-depth interviews with employees of Swiss SMEs that are involved in either purchase processes or day-to-day operations with industrial robots. 

\subsection{Participants}\label{subsec3.1}
Participants were recruited from two Swiss SMEs, both operating in the metal-cutting industry. This sample has been selected since the metal-cutting industry currently sees strong adoption of robots and cobots and therefore offers a very good potential for investigations of actual HRI and related links to OBB. Also, companies in Switzerland are characterized by high personnel expenses, but also face high price levels of goods. They therefore have the option to use robots and can actively decide to do so, while deploying collaborative robots might be less crucial for companies in low-wage countries. In addition, deciders in small companies are still directly involved in purchasing, even though being involved on a strategic level. Regarding sample size, we aim to achieve thematic saturation which means ``the point in data collection when no additional issues or insights are identified and data begin to repeat so that further data collection is redundant, signifying that an adequate sample size is reached''~\cite{HENNINK2022114523}. In their review of 23 publications assessing saturation,~\cite{HENNINK2022114523} report that saturation was reached between 9 and 17 interviews with a mean of 12-13 interviews. This is in-line with one of the first studies empirically assessing saturation indicating that in a homogeneous sample, 6 interviews already capture 80\% of the themes and 12 interviews higher degrees of saturation~\cite{Guest}. Regarding sample homogeneity, all interviewed SMEs use industrial robots as well as cobots to load and unload parts to/from tooling machines and therefore work in the same industry with similar tasks, and in a similar regional area. This homogeneity strengthens the explanatory power and generalizability of our findings for the described sample. In addition, the investigated application represents a commonly used pick-and-place task, which permits us to generalize our results from metal-cutting to many other industries. Table~\ref{tab:tab1} gives an overview of the interviewees. The robots used in the interviewees' companies are several collaborative robots from UR \footnote{See \url{https://www.universal-robots.com/products/}} as well as different articulated Fanuc and Kuka robots \footnote{See \url{https://www.fanuc.eu/si/en/robots/robot-filter-page} and \url{https://www.kuka.com/en-ch/products/robotics-systems/industrial-robots}}. All management and board members (the ``deciders'') were involved in several purchase decisions of industrial robots and all operators have been working with several robots on a daily basis for years (see Table \ref{tab:tab1}). 

\begin{table*}
\caption{Participants Description}
\label{tab:tab1}
\begin{adjustbox}{max width=\textwidth}  
\begin{tabular}{lllllll}

\toprule
\textbf{} & \textbf{Gender} & \textbf{Age} & \textbf{Role in the Company}         & \textbf{Buying Center Role} & \textbf{Highest Education}   & \textbf{Years in Industry} \\ \hline
1         & Male            & 30-40        & Managing Director                    & Decider                     & Polytechnician               & 15+                        \\
2         & Male            & 60-70        & Supervisory Board   Member           & Decider                     & Polytechnician               & 30+                        \\
3         & Male            & 30-40        & Managing director                    & Decider                     & M.A. Business Administration & 5+                         \\
4         & Male            & 40-50        & Machine Operator                     & Operator                    & Basic school                 & 15+                        \\
5         & Male            & 40-50        & Workshop foreman                     & Operator                    & Polytechnician               & 20+                        \\
6         & Male            & 30-40        & Machine Operator                     & Operator                    & Polytechnician               & 15+                        \\
7         & Male            & 40-50        & Plant Manager                        & Decider                     & B.A. Process Engineering      & 15+                        \\
8         & Male            & 20-30        & Machine Operator                     & Operator                    & Polytechnician               & 5+                         \\
9         & Male            & 20-30        & Machine Operator                     & Operator                    & Polytechnician               & 5+                         \\
10        & Male            & 30-40        & Automation \& Digitalisation Officer & Decider                     & M.A. Mechatronics            & 10+                        \\
11        & Male            & 50-60        & Machine Operator                     & Operator                    & Polytechnician               & 30+                        \\
12        & Male            & 50-60        & Machine Operator                     & Operator                    & Polytechnician               & 20+                        \\
13        & Male            & 40-50        & Managing Director                    & Decider                     & Polytechnician               & 25+                        \\
14        & Male            & 30-40        & Machine Operator                     & Operator                    & Polytechnician               & 20+                       
\end{tabular}
\end{adjustbox}
\end{table*}

\subsection{Interview Methodology}
To investigate drivers and barriers from different perspectives, we use the soft laddering interview technique \citep{Gutman,miles2004}. Originating from marketing, laddering is used to understand individuals' behaviors, opinions, attitudes, and beliefs. As laddering offers great value for consumer-focused development of strategies and products \citep{veludo}, it appropriately supports our goal of uncovering role-specific drivers and barriers in purchase and usage situations of robots. Based on means-end theory, laddering enables to uncover not only relevant attributes (A) of a product or a service, but also functional (FC) and psychosocial consequences (PSC) and higher-order personal values (V) that are believed to drive individual behavior~\citep{phillips}. According to this model, consumers prefer products that have attributes which they expect to lead to desired consequences (or prevent undesired consequences), determined by values that are important to them \citep{miles2004}, representing chains from attributes to values (so-called \emph{ACV Chains}). As an example, consider the purchase of an electric vehicle (EV). Several physical attributes of the EV (for instance maximum speed, special car paints, warranty terms, or fuel consumption) can produce functional and psychosocial use-consequences, such as: to get from A to B, travel safely, enjoy the acceleration, avoid fossil fuel consumption or convey a certain image to others. The individual's values (for instance \emph{creativity}, \emph{conservatism}, or \emph{sense of belonging}) lead to an evaluation and rating of the consequences produced by the attributes of the EV, resulting in a decision to use or purchase a certain vehicle. To gain insights on different abstraction levels complies with our goal of achieving a deep understanding of individuals' drivers and barriers when purchasing or using industrial robots.

In the present study, all interviewees have already used or purchased industrial robots. Therefore, we were able to assume substantial experience and well-founded preferences with respect to the \emph{industrial robots} product category during the interviews. The average 15+ years of industry experience strengthens the validity of our results, despite the rather small sample size. To identify important attributes, we started the interview with an approach inspired by \emph{triadic sorting} and \emph{direct elicitation} \citep{BECHLARSEN1999315}, showing the interviewees pictures of a human, an industrial robot, and a cobot, all of them loading parts into a tooling machine. For each of these three ``product alternatives'', interviewees were asked about important attributes on which these alternatives are alike, or rather different from the others. We then repeatedly asked simple queries like ``Why is that important to you?'' to ascend ladders from attributes to consequences and values. All attributes were probed until the ACV chains were exhausted \citep{miles2004}. 

\subsection{Interview and Data Analysis Procedures}

All interviews were conducted in German, in person, took 25-40 minutes each, and were recorded with the consent of the interview partners. As our goal is to investigate drivers and barriers of industrial robots from the interviewees' perspective, we left our questions as open as possible. In connection with open interviews, several classic concerns have to be considered \citep{Price}. To minimize the researcher's power to ``direct, lead or shape the interview''~\citep{Price}, we avoided asking any leading questions and let the respondents answer as freely as possible. Using exactly the same method and process for all interviews allows us to work out and compare differences between the roles interviewed. In addition, asking all participants the same starting questions ensures that they independently highlight those aspects that they value most, and only these aspects are inquired further. We then transcribed all interviews in German, using simple transcription, transcribing word-for-word but slightly smoothed details like leaving out stuttering, word breaks, and hesitation sounds. Transcripts were then coded with MaxQDA Software\footnote{VERBI Software, 2022, Version 22.2.1, \url{https://www.maxqda.com/}} and using Attributes, Functional Consequences, Psychosocial Consequences and Values as main codes, representing the basic structure of ACV chains. In addition, we used content-related sub-codes resulting in 24 relevant attributes, 22 functional consequences, 19 psychosocial consequences, and 14 values. To ensure independence from the author's personal interpretation, several iterations and discussions with a second researcher were carried out until a common interpretation of the codes and the data was achieved. This corresponds to \emph{reliability checks} as described in~\citep{Reynolds2001LadderingTM}. In this process, the number of codes was reduced to 19 final attributes, 15 functional consequences, 13 psychosocial consequences, and 11 values. Table~\ref{tab:RelElements} gives an overview of all relevant codes (in the following collectively referred to as \emph{elements}) per category (A/FC/PSC/V) that were included in the subsequent creation of Hierarchical Value Maps (HVM). For each element, Table~\ref{tab:RelElements} furthermore shows the percentage of respondents that mentioned it, aggregated by group (deciders/operators). Each element was only counted once per respondent, also if it was mentioned several times, and hence this indicates the relevance of the topic for the organizational role. Regarding saturation assessment, 80\% of the most relevant codes across all categories were already mentioned in the third interview with deciders and in the fourth interview with operators, and codes frequency counts indicate sufficient sample size as interviews 9-14 yielded no additional codes. In line with~\cite{Guest,TRAN201771} and given our experience in the field, and with its homogeneity, we argue that this is satisfactory indication that sufficient coverage is reached.

To analyze and further condense the data, means-end chains focus on the relationships between the relevant elements rather than the elements themselves~\citep{Reynolds2001LadderingTM}. To create HVMs, we built ladders for each interviewee's responses, based on mentioned links between the elements. The resulting matrix reflects ``an aggregate map of relationships among elements''~\citep{Reynolds2001LadderingTM}. All relationships among single elements were counted only once for each respondent and, to compare \emph{deciders} and \emph{operators} in our data, we constructed HVMs for each group separately. A cutoff level allows to focus on the most important relationships. Regarding our data and the rather small sample size of our study, we have tested cutoff levels of 2, 3, and 4, indicating that the relationships between two elements represented in the final HVMs have been mentioned at least 2, 3, or 4 times. To ensure sufficient informative value and yet focus on the most important relationships, a cutoff level of 3 was selected, resulting in the inclusion of 40.51\% of all relationships mentioned in total. These remaining relationships are the basis for the construction of HVMs that reflect all connections above the cutoff level. As a consequence of this aggregation, the resulting chains were not necessarily described by the same respondent from attributes to values entirely, but all relationships between elements have been mentioned at least three times. The results are shown in Figures~\ref{fig:Decider-HVM} and \ref{fig:Operator-HVM}. 



\begin{table*}
\caption{Deciders' and Operators' Relevance and Differences of Attributes, Functional Consequences, Psychosocial Consequences and Values}
\label{tab:RelElements}
\begin{adjustbox}{max width=\textwidth}  
\begin{tabular}{p{4cm}p{1.1cm}p{1.2cm}p{0.9cm}p{0.3cm}p{4cm}p{1.1cm}p{1.2cm}p{0.9cm}}
\textbf{Attributes}                                    & Deciders (\%)                 & Operators (\%)               & Difference &  & \textbf{Functional Consequences}                          & Deciders (\%)                & Operators (\%)               & Difference \\ \cline{1-4} \cline{6-9} 
Longevity                                     & \cellcolor[HTML]{FFEB84}66.7  & \cellcolor[HTML]{F8696B}0    & 66.7       &  & Sound Investment                                 & \cellcolor[HTML]{98CE7F}66.7 & \cellcolor[HTML]{F8696B}0    & 66.7       \\ \cline{1-4} \cline{6-9} 
Well-Known   Brand                            & \cellcolor[HTML]{FFEB84}66.7  & \cellcolor[HTML]{FBAA77}12.5 & 54.2       &  & Cost Reduction                                   & \cellcolor[HTML]{63BE7B}83.3 & \cellcolor[HTML]{CCDD82}50   & 33.3       \\ \cline{1-4} \cline{6-9} 
Current Costs                                 & \cellcolor[HTML]{FDCA7D}50.0  & \cellcolor[HTML]{F8696B}0    & 50.0       &  & Image of the Company                             & \cellcolor[HTML]{FFEB84}33.3 & \cellcolor[HTML]{F8696B}0    & 33.3       \\ \cline{1-4} \cline{6-9} 
Operational   Speed                           & \cellcolor[HTML]{FFEB84}66.7  & \cellcolor[HTML]{FFEB84}25   & 41.7       &  & Possible Part Spectrum                           & \cellcolor[HTML]{FFEB84}33.3 & \cellcolor[HTML]{F8696B}0    & 33.3       \\ \cline{1-4} \cline{6-9} 
Unmanned   Operation / Capacity Utilization   & \cellcolor[HTML]{63BE7B}100.0 & \cellcolor[HTML]{A2D07F}62.5 & 37.5       &  & Employee Safety                                  & \cellcolor[HTML]{98CE7F}66.7 & \cellcolor[HTML]{E5E483}37.5 & 29.2       \\ \cline{1-4} \cline{6-9} 
Provider’s   Reaction Time                    & \cellcolor[HTML]{FDCA7D}50.0  & \cellcolor[HTML]{FBAA77}12.5 & 37.5       &  & Increased Output / Sales                         & \cellcolor[HTML]{63BE7B}83.3 & \cellcolor[HTML]{B1D580}62.5 & 20.8       \\ \cline{1-4} \cline{6-9} 
Space   Requirements through Safety Equipment & \cellcolor[HTML]{B1D580}83.3  & \cellcolor[HTML]{C1DA81}50   & 33.3       &  & Adherence to Delivery Dates                      & \cellcolor[HTML]{FFEB84}33.3 & \cellcolor[HTML]{FFEB84}25   & 8.3        \\ \cline{1-4} \cline{6-9} 
Load Capacity                                 & \cellcolor[HTML]{FFEB84}66.7  & \cellcolor[HTML]{E0E383}37.5 & 29.2       &  & Nice to Look at                                  & \cellcolor[HTML]{FFEB84}33.3 & \cellcolor[HTML]{FFEB84}25   & 8.3        \\ \cline{1-4} \cline{6-9} 
Purchase Price                                & \cellcolor[HTML]{FBAA77}33.3  & \cellcolor[HTML]{FBAA77}12.5 & 20.8       &  & Independence of Individual Employees             & \cellcolor[HTML]{FBAA77}16.7 & \cellcolor[HTML]{FFEB84}25   & -8.3       \\ \cline{1-4} \cline{6-9} 
Operating   Range                             & \cellcolor[HTML]{FBAA77}33.3  & \cellcolor[HTML]{FBAA77}12.5 & 20.8       &  & Prevention of Noise                              & \cellcolor[HTML]{F8696B}0.0  & \cellcolor[HTML]{FBAA77}12.5 & -12.5      \\ \cline{1-4} \cline{6-9} 
Flexible   Application                        & \cellcolor[HTML]{FFEB84}66.7  & \cellcolor[HTML]{C1DA81}50   & 16.7       &  & Save Space                                       & \cellcolor[HTML]{F8696B}0.0  & \cellcolor[HTML]{FBAA77}12.5 & -12.5      \\ \cline{1-4} \cline{6-9} 
Safety                                        & \cellcolor[HTML]{FDCA7D}50.0  & \cellcolor[HTML]{E0E383}37.5 & 12.5       &  & Reduce Repetitive, Hard Work                     & \cellcolor[HTML]{63BE7B}83.3 & \cellcolor[HTML]{63BE7B}100  & -16.7      \\ \cline{1-4} \cline{6-9} 
Does What We   Tell It / Predictability       & \cellcolor[HTML]{FFEB84}66.7  & \cellcolor[HTML]{A2D07F}62.5 & 4.2        &  & Avoid Mistakes / Waste                           & \cellcolor[HTML]{98CE7F}66.7 & \cellcolor[HTML]{7DC67D}87.5 & -20.8      \\ \cline{1-4} \cline{6-9} 
Repeat   Accuracy                             & \cellcolor[HTML]{FFEB84}66.7  & \cellcolor[HTML]{A2D07F}62.5 & 4.2        &  & Save Time                                        & \cellcolor[HTML]{F8696B}0.0  & \cellcolor[HTML]{FFEB84}25   & -25.0      \\ \cline{1-4} \cline{6-9} 
Simple   Programming                          & \cellcolor[HTML]{B1D580}83.3  & \cellcolor[HTML]{63BE7B}87.5 & -4.2       &  & Planning Security                                & \cellcolor[HTML]{F8696B}0.0  & \cellcolor[HTML]{B1D580}62.5 & -62.5      \\ \cline{1-4} \cline{6-9} 
Stability                                     & \cellcolor[HTML]{F98971}16.7  & \cellcolor[HTML]{FFEB84}25   & -8.3       &  &                                                  &                              &                              &            \\ \cline{1-4}
Human   Supremacy                             & \cellcolor[HTML]{FBAA77}33.3  & \cellcolor[HTML]{C1DA81}50   & -16.7      &  & \textbf{Values}                                           & Deciders (\%)                & Operators (\%)               & Difference \\ \cline{1-4} \cline{6-9} 
Design                                        & \cellcolor[HTML]{F8696B}0.0   & \cellcolor[HTML]{FFEB84}25   & -25.0      &  & Take Responsibility                              & \cellcolor[HTML]{63BE7B}83.3 & \cellcolor[HTML]{F8696B}12.5 & 70.8       \\ \cline{1-4} \cline{6-9} 
Moves Quietly / Stops Smoothly / Dynamic      & \cellcolor[HTML]{F8696B}0.0   & \cellcolor[HTML]{FFEB84}25   & -25.0      &  & Long-Term Thinking / Karma                       & \cellcolor[HTML]{B2D580}66.7 & \cellcolor[HTML]{F8696B}12.5 & 54.2       \\ \cline{1-4} \cline{6-9} 
                                              &                               &                              &            &  & Self-Realization / Commitment to   Quality       & \cellcolor[HTML]{63BE7B}83.3 & \cellcolor[HTML]{FFEB84}62.5 & 20.8       \\ \cline{6-9} 
\textbf{Psychosocial Consequences}                     & Deciders (\%)                 & Operators (\%)               & Difference &  & Financial Autonomy                               & \cellcolor[HTML]{63BE7B}83.3 & \cellcolor[HTML]{63BE7B}87.5 & -4.2       \\ \cline{1-4} \cline{6-9} 
Social Responsibility for   Employees         & \cellcolor[HTML]{98CE7F}66.7  & \cellcolor[HTML]{F8696B}12.5 & 54.2       &  & Collegiality / Cooperation                       & \cellcolor[HTML]{FFEB84}50.0 & \cellcolor[HTML]{FFEB84}62.5 & -12.5      \\ \cline{1-4} \cline{6-9} 
Employee   Satisfaction                       & \cellcolor[HTML]{63BE7B}83.3  & \cellcolor[HTML]{CCDD82}75   & 8.3        &  & Forward-Looking Attitude / Learn   Something New & \cellcolor[HTML]{FFEB84}50.0 & \cellcolor[HTML]{FFEB84}62.5 & -12.5      \\ \cline{1-4} \cline{6-9} 
Future Safety   / Livelihood                  & \cellcolor[HTML]{63BE7B}83.3  & \cellcolor[HTML]{CCDD82}75   & 8.3        &  & Orderliness                                      & \cellcolor[HTML]{F8696B}0.0  & \cellcolor[HTML]{F8696B}12.5 & -12.5      \\ \cline{1-4} \cline{6-9} 
Strengthen the Domestic Economy               & \cellcolor[HTML]{FFEB84}33.3  & \cellcolor[HTML]{F98971}25   & 8.3        &  & Loyalty                                          & \cellcolor[HTML]{F8696B}0.0  & \cellcolor[HTML]{F98971}25   & -25.0      \\ \cline{1-4} \cline{6-9} 
Avoid Risks                                   & \cellcolor[HTML]{FBAA77}16.7  & \cellcolor[HTML]{F8696B}12.5 & 4.2        &  & Health                                           & \cellcolor[HTML]{F8696B}0.0  & \cellcolor[HTML]{FBAA77}37.5 & -37.5      \\ \cline{1-4} \cline{6-9} 
Social   Connections                          & \cellcolor[HTML]{CCDD82}50.0  & \cellcolor[HTML]{FFEB84}62.5 & -12.5      &  & Peace of Mind                                    & \cellcolor[HTML]{FCBF7B}33.3 & \cellcolor[HTML]{63BE7B}87.5 & -54.2      \\ \cline{1-4} \cline{6-9} 
Pleasure \&   Fascination of Technology       & \cellcolor[HTML]{98CE7F}66.7  & \cellcolor[HTML]{98CE7F}87.5 & -20.8      &  & Enthusiasm                                       & \cellcolor[HTML]{FA9473}16.7 & \cellcolor[HTML]{B1D580}75   & -58.3      \\ \cline{1-4} \cline{6-9} 
Satisfaction   of Management                  & \cellcolor[HTML]{FBAA77}16.7  & \cellcolor[HTML]{FDCA7D}50   & -33.3      &  &                                                  &                              &                              &            \\ \cline{1-4}
Pursue a   Meaningful Activity                & \cellcolor[HTML]{98CE7F}66.7  & \cellcolor[HTML]{63BE7B}100  & -33.3      &  &                                                  &                              &                              &            \\ \cline{1-4}
Aesthetics                                    & \cellcolor[HTML]{F8696B}0.0   & \cellcolor[HTML]{FBAA77}37.5 & -37.5      &  &                                                  &                              &                              &            \\ \cline{1-4}
Fear of Job   Loss                            & \cellcolor[HTML]{FFEB84}33.3  & \cellcolor[HTML]{CCDD82}75   & -41.7      &  &                                                  &                              &                              &            \\ \cline{1-4}
Customer   Satisfaction                       & \cellcolor[HTML]{FBAA77}16.7  & \cellcolor[HTML]{FFEB84}62.5 & -45.8      &  &                                                  &                              &                              &            \\ \cline{1-4}
Keep Quality   Promises                       & \cellcolor[HTML]{FBAA77}16.7  & \cellcolor[HTML]{CCDD82}75   & -58.3      &  &                                                  &                              &                              &            \\ \cline{1-4}

\end{tabular}
\end{adjustbox}
\end{table*}

\section{Results}
\label{sec:results}
\begin{figure*}[ht]
    \centering
    \includegraphics[width=1.0\textwidth]{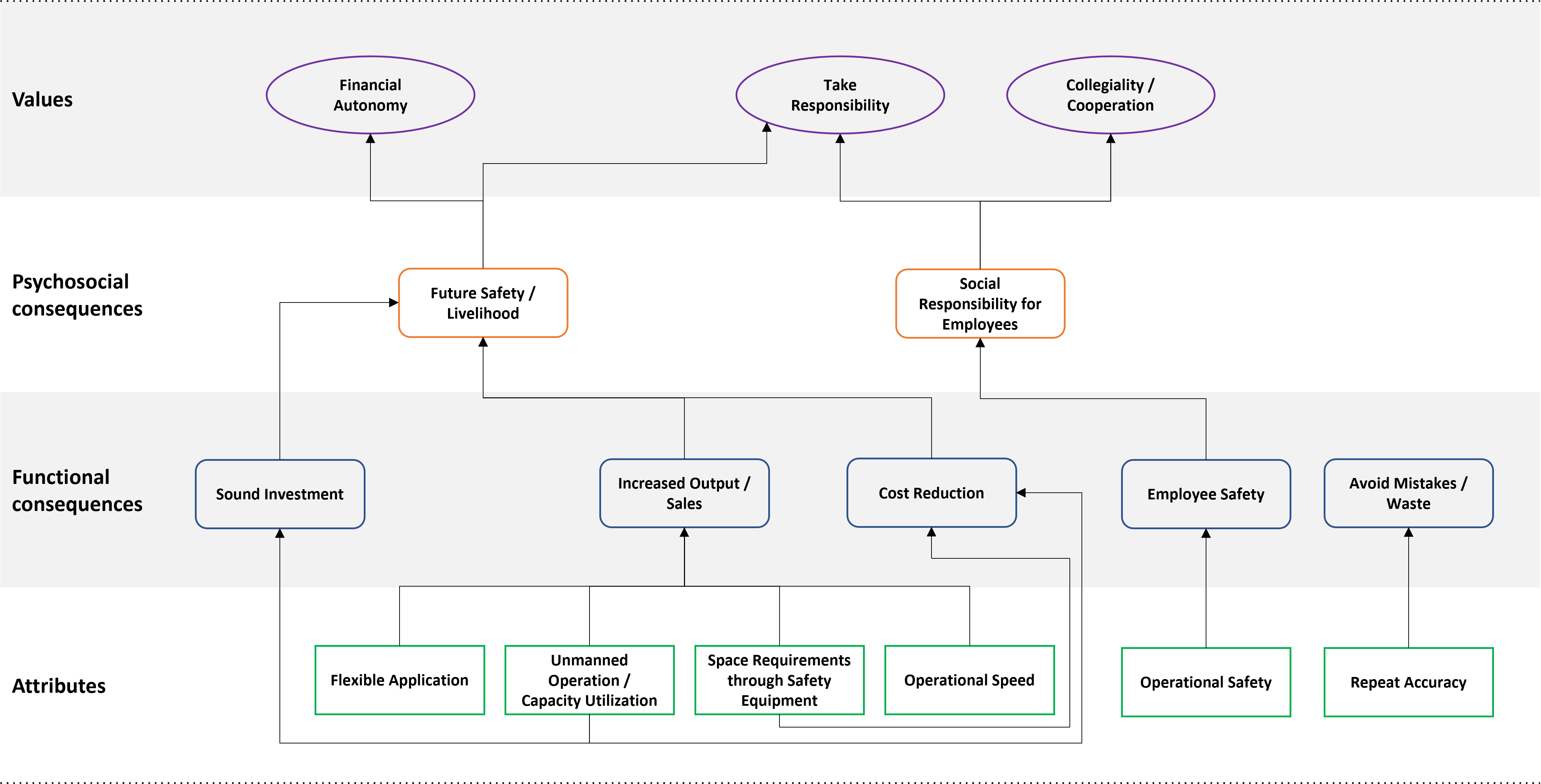}
    \caption{Deciders' Hierarchical Value Map}
    \label{fig:Decider-HVM}

    \vspace{10mm}

    \includegraphics[width=1.0\textwidth]{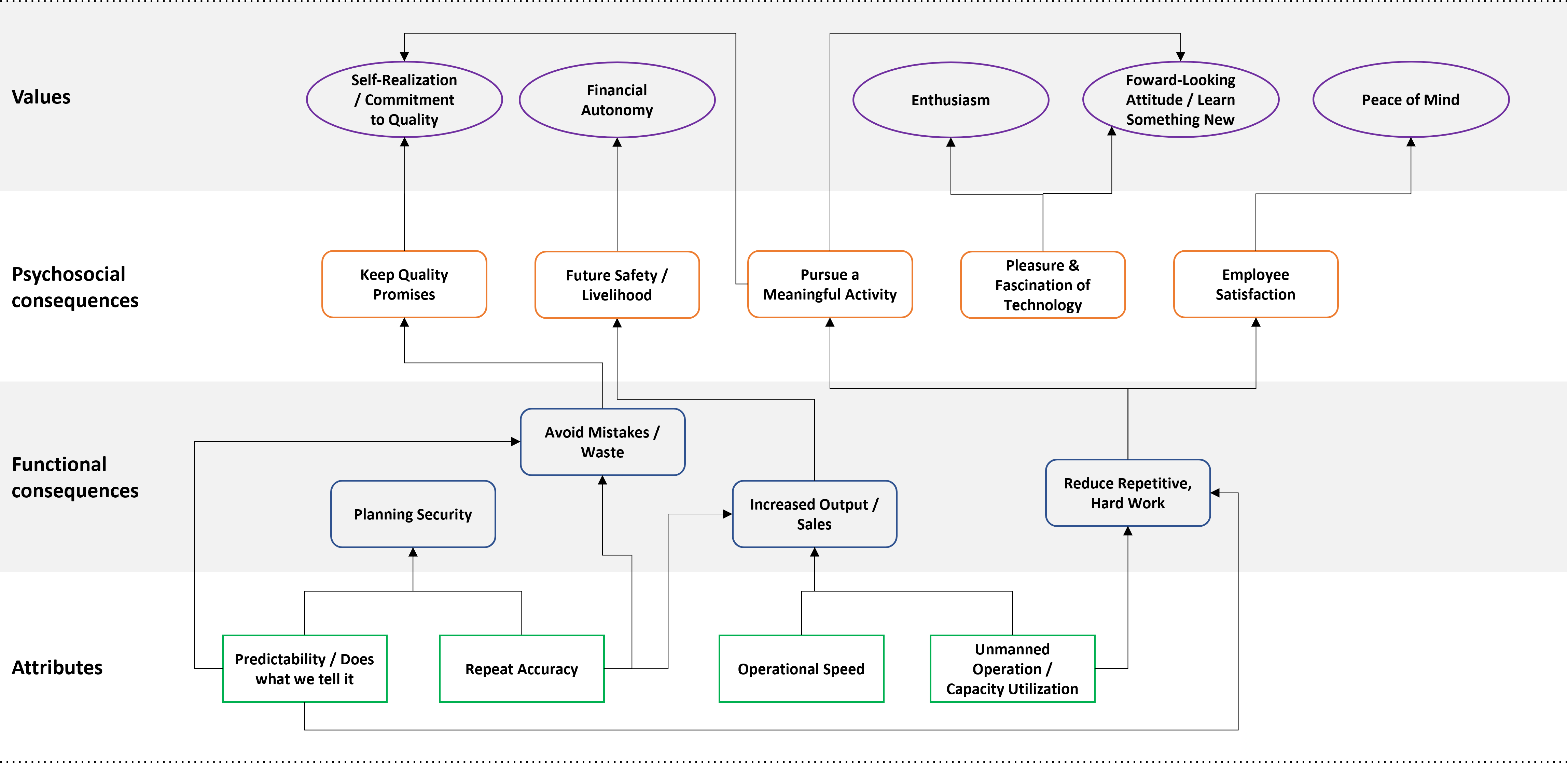}
    \caption{Operators' Hierarchical Value Map}
    \label{fig:Operator-HVM}
\end{figure*}

Deciders deal with a different task when purchasing a collaborative robot compared to operators who work together with robots in their daily lifes, and operators accordingly seem to associate robots with more operational and emotional topics. Our results reflect this on all abstraction levels, such as that the representation of psychosocial consequences and values are much broader, evident in the operators' HVM. Deciders seem to be involved rather functionally and focused on their organizational roles, even though there are many similarities on all abstraction levels that drive both groups' behaviors and attitudes. To identify similarities and differences between deciders and operators, our results are analysed and presented as follows:

\begin{itemize}
    \item Table~\ref{tab:RelElements} shows all elements across A, FC, PSC and V as well as the relevance per element (Columns 2 \& 3 in each table), green indicating high relative relevance. Column 4 in Table~\ref{tab:RelElements} presents the differences between the two groups, sorted by large differences indicating what was frequently mentioned by deciders but not operators (top), and vice versa (bottom).
    \item Figures~\ref{fig:Decider-HVM} and \ref{fig:Operator-HVM} illustrate how both groups connect these elements in form of the HVMs.
    \item In addition to these quantitative results, we in the following discuss a selection of quotes to illustrate how both groups discuss the topics in their own words. 
\end{itemize}

On the attribute-level, \emph{unmanned operation} is most important for deciders, \emph{space requirements through safety equipment} and \emph{simple programming} being second most important. Operators rate \emph{simple programming} to be most important, followed by \emph{unmanned operation}, \emph{predictability}, and \emph{repeat accuracy}. Other major differences include that \emph{robot brand}, \emph{longevity} of the robot and \emph{current costs} are relevant for \emph{deciders only}, whereas \emph{design} and \emph{smooth operation} are mentioned by \emph{operators only}. Even though some elements are similarly important for both groups, deciders seem to value rather strategic, high-level topics, whereas operators consider elements that directly affect their daily operations to be important. With regard to the HVMs and relationships between attributes and functional consequences, the divergences become even more visible. Deciders mostly link their relevant attributes to consequences related to the company's success, whereas operators see additional benefits that simplify their work routine. Regarding \emph{unmanned operation}, quotes from both groups support these different views: Operator 11 mentions \emph{``He [the robot] then does the work. And I have to lift around less. I also prefer
programming, that’s more interesting, you need your brain for it.''} Juxtaposed, Decider 3 mentions: \emph{``In the end, this is lucrative and saves money. A robot costs a one-off amount and pays for itself. A worker continually costs money.''}

The robots' attributes lead to different consequences on the functional and psychosocial levels. Both groups consider the \emph{reduction of repetitive, hard work} favorable functional consequences of industrial robots. However, deciders value \emph{cost reduction} and \emph{increased output} equally important, while \emph{avoiding mistakes and waste} is second most important for operators. \emph{Making sound investments} is important for deciders but not for operators, and \emph{planning security} as a functional consequence of the attributes \emph{predictability} and \emph{repeat accuracy} only seems to be relevant for operators, demonstrating another topic that directly affects their daily work routine. Moreover, the operators' HVM demonstrates a detailed structure from the causative attributes \emph{unmanned operation} and \emph{predictability} to the functional consequence \emph{reduction of repetitive, hard work}, to the associated pyschosocial consequences \emph{pursuing meaningful activities} and \emph{employee satisfaction}, whereas deciders link \emph{unmanned operation} to the functional consequences \emph{making sound investments}, \emph{increased output} and \emph{cost reduction} which are all linked to the psychosocial consequence \emph{future safety}.

Regarding psychosocial consequences, \emph{employee satisfaction} and \emph{future safety} are considered most important for deciders, and \emph{pursuing a meaningful activity} is most important for operators. A major difference, which might be in the nature of things, is that deciders shoulder \emph{social responsibility for their employees}, arising from the functional consequence \emph{employee safety} and the attribute \emph{operational safety}. \emph{customer satisfaction} and \emph{to keep quality promises} are considered important mostly for operators.

These explanations on the consequence-level create an interesting picture of differing responsibilities and results when looking at deciders' and operators' evaluations of robots. The following quotes describe some of the above-mentioned differences in the interviewees' own words: Operator 8 mentions \emph{``It only does what I tell him. That's good, he has to do it. He can't do what he wants. When I give him a task he has to do that and nothing else. Otherwise mistakes can happen, a serious crash or something, it's not a toy.''} Contrary to this, Decider 7 mentions \emph{``Thanks to robots, we earn money, that's how we live. It works 24/7 and doesn't need breaks, our machines have to run and produce chips.''}
 
Unsurprisingly given the work context, both groups are driven by the value of \emph{financial autonomy}. Also, the relevance of \emph{taking responsibility} and \emph{long-term thinking}, linked to the own \emph{future safety} as well as those of the employees is a logical result of the deciders' roles and represents a rather rational mindset. 
On the other hand, the connections between consequences and values seem to be much more layered for operators, represented by the values of \emph{peace of mind}, corresponding to the advantages of robots for their operational, daily work, and \emph{enthusiasm}, linked to the psychosocial consequence of \emph{fascination of technology}. In addition, operators link working with robots through the psychosocial consequence of \emph{pursuing a meaningful activity} to the values of \emph{self-realization} and to their \emph{forward-looking attitude}.

Even though most topics emerge from the analysis above, two topics that are discussed often in both, research and practice, deserve to be mentioned separately: \emph{fear of job loss} and \emph{safety}. The potential of robots replacing jobs was mentioned by 75\% of the operators and by 33.3\% of the deciders. Even though some parts of today's tasks in the metal-cutting industry will be likely taken over by robots, the necessity of deploying robots was not questioned by either group. The following quotes demonstrate this: \emph{``Robots are taking away jobs, especially simple jobs. They won't be able to do everything, but that can cause anxiety. It doesn't bother me because I do a lot of work that a robot can't do. He cannot think for himself.''} (Operator 4), and \emph{``Our people just have to work with it. If we don't develop further and increase productivity and generate growth, then the future will be limited or at risk''} (Decider 7).

Safety-related concerns were mentioned by some deciders (50\% on the attribute-level, 66.7\% on the functional consequence-level), but less so by operators (37.5\% on both, attribute- and functional consequence-level). However, this does not reflect low importance of the topic, since robotic cells and cobots are required to meet certain safety standards by law in Switzerland. Therefore, deciders consider this standard when buying robots, and operators feel safe working with the robotic devices installed in their shopfloors. Operator 14 mentions \emph{``If the basic requirements are met, not much can happen. I have no concerns at all about working with these systems. You have to know what to do and which button to press, then it will work safely''} and Decider 1 mentions \emph{``The whole thing has to be safe, that's a standard that we don't question. That's why it's not a big issue, it's been in the concept from the start, it's not a point of discussion with suppliers either, it's covered. There is no cost/benefit consideration, there is only 100\%''} (Decider 1).

For interested readers, a table of supplementary quotes can be found in the Appendix, providing additional statements in the deciders' and operators' words.

\section{Discussion}
\label{sec:discussion}
The present study qualitatively investigates drivers and barriers of industrial robots and cobots in Swiss SMEs, comparing deciders' and operators' perspectives. As both groups are involved in different organizational tasks, we compare two HRI contexts: the purchase decision process in an SME and daily usage of robots to load and unload tooling machines in manufacturing shopfloors. As deciders are mainly involved in purchase decisions and operators mainly work with robots on a daily basis, their ratings of important aspects as well as their judgements differ. In general, operators consider a variety of topics and seem to be affected more emotionally, whereas deciders seem to be more rationally involved in their robotic-related tasks. With regard to drivers and barriers, the topics presented in Section~\ref{sec:results} can be seen as drivers whenever the discussed attributes lead to the desired consequences (e.g., if the robot is predictable, this reduces repetitive hard work, and thus increases employee satisfaction), and as barriers whenever certain attributes do not exist (e.g., inability to meet quality promises if missing predictability causes waste). \citet{Cigdem} found significant effects of negative attitude towards robots and trust in robots on operators' intention to use them. The companies interviewed in our study already use industrial robots, and all operators consistently expressed positive attitudes towards robots, even though we cannot make a statement regarding their intention to use robots before robots were installed. Furthermore, our findings cannot confirm the effect of trust on intention to use; however, our findings indicate that trust in robots during operation depends on several concrete attributes such as predictability and repeat accuracy. In~\citet{Kopp2020}, occupational safety was rated most important by decision-makers in the German manufacturing industry. In addition, fear of job loss was rated important, and trusting a cobot was rated the most important human factor when introducing cobots. Further important factors were appropriate cobot configuration (such as suitability of the whole production process, task allocation and positioning of working materials), and financial factors, whereby operational costs were rated more important than one-time acquisition and maintenance costs. Our results support the importance of all these findings, even though the weighting and assessments partially differ. Regarding safety, the respondents in our study confirm the importance, however, legal requirements not only reduce the importance of this topic for deciders, but also operators did not state any reservations. As illustrated in Section~\ref{sec:results}, fear of job loss was not considered a barrier for the operators interviewed in our study. However, this finding might be biased by the rather high educational level of all respondents and their supremacy in case of creative, more challenging tasks like programming and optimization. Even though they do not fear that robots replace their jobs, they are aware that unskilled workers performing simple tasks might be replaced by robots. Similar to the explanations above, trusting robots was not mentioned literally in our study. However, operators consider \emph{predictability} -- that the robot does what we tell it -- and \emph{repeat accuracy} as important, representing the basis for trust in what the robot does by enabling several favourable consequences. From the deciders' points of view, these two attributes were important as well, however, they linked \emph{repeat accuracy} to \emph{avoiding mistakes} only, supporting that they evaluate robots more rationally. Our study focuses exclusively on the task of loading/unloading tooling machines, and appropriate configuration of the robot was therefore not discussed explicitly, probably because all robotic systems installed in the interviewed companies were designed and built for these applications, and hence the configuration was not questioned. Financial factors were, not surprisingly, mostly mentioned by deciders. Similar to~\citet{Kopp2020}, current costs of operation were mentioned slightly more often than one-time purchase prices. However, the cost of the robots themselves seem to be of secondary importance while increased output, cost reduction and making sound investments were relevant consequences mostly for deciders, and resulting future safety was linked to financial autonomy by both, deciders and operators.

With regard to task and non-task dimensions in organizational decision-making, both dimensions have been identified for both groups. For example, deciders not only take responsibility for the future safety of the company, but also with respect to the employees and their families, where they link both to the decision of deploying robots in manufacturing shopfloors. Operators, on the other hand, do not only appreciate the relief of monotonous work, but also the possibility to learn new things. This is also reflected in several psychosocial consequences that drive particularly operators' evaluation and behavior when working with robots. Differentiating between the two roles investigated might play a relevant role for certain needs, wishes, and reservations when optimizing interactions or designing more acceptable industrial robots, relating to possible moderators presented in Section~\ref{sec2}. Moreover, laddering and means-end theory enables to deepen the understanding of what robot attributes lead to acceptance or rejection of robot purchase and usage and shed light on how these lead to formation of an opinion on different abstraction levels. The inclusion of the investigated organizational factors -- roles and context -- enriches the understanding of what promotes or hinders robot deployment, representing the foundation of robot acceptance with regard to robot-, human- and context-specific aspects. With regard to the challenge of integrating marketing and Supply Chain Management as demanded in \citet{CABANELAS202365}, we believe that a better understanding of the BC portrayed by different roles and motives involved aligns the robot suppliers' manufacturing oriented focus of purchasing departments with the marketing departments' aim to fulfill customer needs. In addition, consideration of such findings not only supports robot suppliers to meet customer needs, but also whole economies by increasing automation states and thereby to stay competitive in the long-run.

\subsection{Managerial Implications}

A deep understanding of what drives and hinders robot acceptance has the potential to inform suppliers as well as demanders of industrial robots and builds the foundation for robot deployment as well as robot acceptance. Recent findings in the hospitality industry suggest managers to not only advertise robots' advantages, but also their limitations, to prevent fear of job-loss, and to ensure successful collaborations between humans and robots~\cite{Chen}. To put this into practice, affected stakeholders require knowledge of what these advantages and limitations are \emph{in the perception of actual robot customers}, and how to create a successful collaboration between robots and humans on this basis. Our findings contribute to this knowledge in three ways. First, we present a list of relevant attributes, consequences and values that constitute robot acceptance and/or rejection. Second, we provide quantified relevancies for these elements, and how they differ between operators and deciders. And third, we demonstrate how the elements are linked with each other on different abstraction levels, and how these links differ between the two groups. These links provide an additional potential for demanders as well as for suppliers. By understanding how attributes are linked to higher-level values, these ladders allow to build segments of individuals that share certain values and moral concepts, but also to understand how they are linked to concrete product attributes that can be used to operationalize these findings. For marketing and sales in B2B, this enriches the data basis towards relevant personal characteristics beyond basic demographics, and these personality profiles are believed to play a relevant role in the purchasing process~\cite{Mier}.

Suppliers of robots are interested in understanding their customers' motives for or against buying a robot, representing the basis for robot deployment and, accordingly, sales. Demanders, on the other hand, are organizations that buy robots and where both, deciders and operators, are interested in achieving robot acceptance to improve their daily operations. Even though deciders have partly similar interests to robot suppliers, namely to ``sell'' robots and to increase operator acceptance, our study points out that relevant factors between robot suppliers and deciders differ. For example, we show that \emph{safety} is relevant for deciders and operators alike, but corresponds to a prerequisite rather than a decision criterion that robot suppliers can use as selling arguments. \emph{Trust}, on the other hand, is mostly required by operators, and the presented HVMs demonstrate that attributes such as the robot's predictability and repeat accuracy are appreciated by operators because they do help them to avoid mistakes, and support operators' will for self-realization. Product presentations for deciders should therefore focus on the presented attributes and resulting consequences that enable achievement of the most desired values \emph{responsibility}, \emph{self-realization} and \emph{financial autonomy} whereas the most desired values for operators that suppliers should focus on are \emph{financial autonomy}, \emph{peace of mind} and \emph{enthusiasm}. 

The organization that intends to buy a robot, on the other hand, is managed by deciders, and our results support them in their efforts to convince operators to use and accept robots. Insights about operators' reasons for and a detailed understanding of their fears helps deciders to create a favourable baseline in the organization.
On the other hand, understanding their customers' perspectives and the different motives involved in a buying decision helps suppliers to meet demands resulting in higher buying intentions, and sales. In addition, by understanding operators' motives, they are not only able to increase customers' willingness to buy but also to support deciders in their challenge when convincing operators that robots support them in reaching financial autonomy, self-realization and peace of mind.

In addition, even product designers benefits from adapting products to the wishes of actual users. As robots are used by operators, and their wishes do not necessarily correspond to those of decision-makers, our results allow finer tuning of product attributes to the actual customer's needs. This study provides this information by pointing out which elements are relevant, how relevant they are, how these relevancies differ per group, how the elements are linked to each other, and how these linkages differ between operators and deciders. On a more abstract level, the grading of customer needs on different abstraction levels allows suppliers as well as demanders to align not only product attributes and customer needs, but also value conceptions. Suppliers can use the relevant values worked out in this study to design marketing campaigns, and demanders can implement a strategy and culture to explicitly link the identified attributes with values, creating a robot acceptance and deployment promoting setting.

\subsection{Limitations}
Even though this study has a clear focus, role- and context-wise, it has several potential limitations. With regard to the two contexts investigated, only operators actually collaborate directly and everyday with industrial robots. Although deciders are involved in robot operations and purchasing of robots, which brings them closer to the topic than most individuals, they do not regularly interact directly with robots and these interactions might therefore not be considered as HRI in the narrow sense. However, purchasing processes are an essential part of the actual occurrence of interactions with robots, since a decision to buy a robot is needed before interactions can emerge at all. With regard to OBB theory, our findings represent a first access to understand BC actors' motives for a specific industrial product, investigated in SMEs that differ from larger organizations with more sophisticated and formal buying processes. However, the existing state of research requires more data from different samples, and we hence argue that our study provides a first effort into such elaborations that should be conducted world-wide, and that one of the contributions of our study is to motivate others to repeat and extend our investigation in a variety of geographical and task settings. In addition, BC structures and decision-making processes when buying industrial robots require additional investigations in future. With regard to HRI research, deciders' opinions are crucial and valuable to be included as the organizational perspective is still mostly ignored. Also, the application of loading and unloading a tooling machine in manufacturing shopfloors is only one application of robots and can be further divided into sub-tasks, enabling additional context-specific findings, enabling better understanding of BCs and OBB. Nevertheless, robots are widely used for pick-and-place tasks in manufacturing shopfloors, hence representing a typical application in industrial practice. In view of the rather small sample size, several limitations apply. The analysed companies operate in a specific industrial domain and in a limited regional environment. Moreover, the robots used in the company's shopfloor only represent three different brands and two types of robots (articulated robots/cobots), and the interviewees' experiences are limited to those two types of robots. Still, our sample provides valuable insights into actual robot users' decision-making structures, as compared to students or novices that represent most of the samples in existing HRI-studies. Regarding the product categories of industrial robots and cobots, all participants have gained experience with both. Therefore, findings might be influenced from one or the other, and cannot be related to industrial robots or cobots exclusively. Finally, the companies included in our study only employ male deciders and operators. While this is representative of the metal-cutting industry in the region, our findings are hence biased.  

\subsection{Conclusions And Future Research}
Our study reveals several differences between deciders and operators involved in robot-related tasks in manufacturing companies. These insights not only help to understand motives and concerns of involved individuals, but also how the deployment of robots in industry actually comes into existence. Despite the above-mentioned limitations, the presented findings enable a variety of follow-up research streams. In view of the lack of moderator-specific results in HRI, OBB allows to focus on organizational roles when designing robot attributes that reinforce acceptance. To generalize our conclusions, we propose to use our findings for future quantitative studies, verifying the relevance of specific topics with a larger, representative sample. Our approach represents a start into such research by gaining deeper understanding of an individual’s organizationally oriented decision-making process in the product categories of industrial robots  and cobots, building the foundation for further extension of the BC and abstraction regarding OBB in the future. Based on our presentation of how the perception and evaluation of attributes are influenced by higher-level consequences and values, product design as well as marketing strategies can be adapted to certain user groups, increasing product-market fit and subsequently, customers' intention to buy and use industrial robots. In addition and with regard to automation and digitalization advancements, detecting individuals and their relevant characteristics such as their roles autonomously, and to enrich these findings with additional sensor data might allow future robot systems to adapt behaviors to individuals at run time, thereby realizing human-awareness and reactivity to specific individual needs. Regarding robot deployment, understanding purchase situations enables manufacturers and suppliers of industrial robots to profit substantially by addressing role-specific needs and reservations with certain robot characteristics, enabling exciting opportunities for future marketing automation. To further explore these interrelations between BC roles and role-specific needs and reservations, additional research is needed with regard to additional roles as well as in different contexts and phases of the purchasing process. We suggest that the inclusion of organizational processes has opened a door towards a more broader understanding of actual robot usage in practice. Matching robot characteristics with organizational aspects such as role-specific requirements might lead not only to further deployment of industrial robots in shopfloors, but also to increased acceptance on individual levels and thereby ensure economies' competitiveness for the future.

\subsection*{Ethical Approval}

The Ethics Committee of the author's university has confirmed that no ethical approval is required.

\bibliographystyle{agsm}
\bibliography{References}

\appendix

\begin{table*}
\label{Supplementary Quotes}
\caption{Supplementary Quotes from Interviewees}
\begin{adjustbox}{max width=\textwidth}  
\begin{tabular}{p{2.7cm}p{7.7cm}p{7.7cm}}

                                                                                                           & Operators                                                                                                                                                                                                                                                                                                                                                     & Deciders                                                                                                                                                                                                                                                                                                                                                                \\ \hline
\multirow{2}{*}{\begin{tabular}[c]{@{}l@{}}More exciting \\ work vs. functional  \\ advantages\end{tabular}} & ``Operator 5: I   like that you can put parts in, press start, then they run through and you   can do other things. The tedious work of putting parts in, starting, taking   them out and so an. There are people who think it's great, I don't find that   interesting.''                                                                                   & ``Decider 1: Of course, the robot is financially interesting because it doesn't take vacation   time, doesn't cause any non-wage labor costs, is very low-maintenance, etc.''                                                                                                                                                                                       \\ \cline{2-3} 
                                                                                                          & ``Operator 8: It takes over the work that I would otherwise   have to do by hand. While the robot loads the machine, I can program and do   other things during runtime. That's more exciting, programming and optimizing   is what I prefer over clamping  parts. That's more exciting,   you don't always do the same thing.''               & ``Decider 10: Productivity   has to be good so that we are able to compete with others who do for example   still do it by hand and if we get three shifts out of it, that's different   than if it only runs from 8-5.''                                                                                                                                             \\ \hline
\multirow{2}{*}{\begin{tabular}[c]{@{}l@{}}Reliability is very \\ important\end{tabular}}                  & ``Operator 2: I   have to rely on the robot. If it wasn't precise enough, I wouldn't use it at   all, it would result in waste. I can't always stand there. Then I'd rather do   it myself, then I don't need the robot.''                                                                                                                                   & ``Decider 2: The   robot is more predictable, you know what is happening. With humans, you pay   them and if they don't pay attention it's more annoying than with robots.   Because they could have noticed it if you were paying attention. With a robot   we know that it can't do that.''                                                                         \\ \cline{2-3} 
                                                                                                           & ``Operator 12: If the robot isn't reliable, that would be   nonsense and it would be unnecessary. If I have to stand next to it and it   doesn't move forward on its own, that's no use. But he doesn't actually do   anything wrong on his own, it's always external things that stop him.''                                                              & ``Decider 3: The   robot is fixed, is set up, and does exactly what you tell it to do. The robot   is stupid and only does what you tell it to do. He doesn't think.''                                                                                                                                                                                                \\ \hline
\multirow{2}{*}{\begin{tabular}[c]{@{}l@{}}Job loss vs. \\ future safety\end{tabular}}                   & ``Operator 11: Fewer   people will be needed, but that doesn't stress me out. There are parts the   robot can't do.''                                                                                                                                                                                                                                        & ``Decider 2: Fear   of job loss? I actually don't take it into account at all. Today it is part   of our industry or the profession of polymechanic. That's the standard. If   someone can't work with it, they're not in the right place with us.''                                                                                                                  \\ \cline{2-3} 
                                                                                                           & ``Operator 9: Luckily the robot can't do everything, if the   robot could do everything it would need fewer employees, that would be bad   for us.''                                                                                                                                                                                                         & ``Decider 3: I   do understand them, but if we don't move forward we'll all end up without a   job.''                                                                                                                                                                                                                                                                 \\ \hline
\multirow{2}{*}{\begin{tabular}[c]{@{}l@{}}Safety: personal \\concern vs. \\ responsibility\end{tabular}}    & ``Operator 6: Because   people can suffer, these are my work colleagues, I don't want anything to   happen to them. If you have heavy parts, you don't want to lift them around   500 times all day or do the same movement, that's physical strain and in the   worst case it leads to illness. I want all colleagues to be healthy until   retirement.''  & ``Decider 3: Injured   employees are a very high cost factor for us and cause health costs. It is   very harmful for the employee, he is in pain, is in the hospital, may not get   well at all, and also drives up the insurance premiums. Industrial accidents   are fundamental to avoid, that's what every entrepreneur is committed to.''                        \\ \cline{2-3} 
                                                                                                           & ``Operator 4: It is important for me that it stops, that it   is collaborative. It happens, when you get too close to it, it stops and work   can simply be resumed. Also that it stops finely. When it started to spin, it   would breaks things.''                                                                                                        & ``Decider 1: Because   I am a responsible person and I see this as an important quality for an   employer.''                                                                                                                                                                                                                                                          \\ \hline
\multirow{2}{*}{\begin{tabular}[c]{@{}l@{}}Common goal to \\ make money\end{tabular}}                      & ``Operator 4: That's   good for us too, the better things go, the better the company is doing and   the more money comes in''                                                                                                                                                                                                                                & ``Decider 13: If we have free capacities, we can train the employees so that higher wages are possible while the robot works and does not need a wage. This will   certainly also produce happier employees with more motivation to continue   their education, and also happier families behind them, and   suddenly there is house in it.''  \\ \cline{2-3} 
                                                                                                           & ``Operator 12: With robots we make more parts, finish them   faster, do more work, and that makes more money. If the company is doing   well, we are doing well. Then I'll be fine and I can be confident that I   won't lose my job. Otherwise I won't make any money.''                                                                                    & ``Decider 10: We   have high wages in Switzerland, if we are not competitive, the parts will be   made somewhere else. So if we want to do them here we have to be more   efficient than others in order to be able to pay the wages.''                                                                                                                              \\ \cline{2-3} 
                                                                                                           & ``Operator 14: The prices have to be right, the customer has   to be satisfied. Then he orders again, we have work, and I can maintain my   standard of living.''                                                                                                                                                                                            & ``Decider 1: Because   they make a good contribution to ensuring that we are doing well in turn. And   because we are responsible for ensuring that they can feed their families in   the long term.''                                                                                                                                                               
\end{tabular}
\end{adjustbox}
\end{table*}

\end{document}